\documentclass[letterpaper, 10 pt, conference]{ieeeconf}
\usepackage{times}
\usepackage[margin=0.77in]{geometry}

\usepackage{multicol}
\usepackage[bookmarks=true]{hyperref}
\usepackage{xcolor}
\usepackage{graphicx}
\usepackage{booktabs}
\usepackage[normalem]{ulem}
\usepackage{amsmath}
\usepackage{dirtytalk}

\newif\ifpaperfinal
\paperfinalfalse
\ifpaperfinal
\newcommand{\stnote}[1]{}
\newcommand{\ngnote}[1]{}
\newcommand{\tnnote}[1]{}
\newcommand{\bbnote}[1]{}
\newcommand{\vcnote}[1]{}
\newcommand{\gdknote}[1]{}
\else
\newcommand{\stnote}[1]{\textcolor{blue}{\textbf{ST: #1}}}
\newcommand{\ngnote}[1]{\textcolor{green}{\textbf{NG: #1}}}
\newcommand{\tnnote}[1]{\textcolor{teal}{\textbf{TN: #1}}}
\newcommand{\bbnote}[1]{\textcolor{red}{\textbf{BB: #1}}}
\newcommand{\gdknote}[1]{\textcolor{orange}{\textbf{GDK: #1}}}
\newcommand{\vcnote}[1]{\textcolor{red}{\textbf{VC: #1}}}
\fi

\begin{document}

\title{\LARGE \bf Grounding Language Attributes to Objects using Bayesian Eigenobjects}


\author{Vanya Cohen*, Benjamin Burchfiel*, Thao Nguyen*, Nakul Gopalan,
    Stefanie Tellex,
    George Konidaris
    \\ vanya\_cohen@alumni.brown.edu, bcburch@cs.duke.edu,\\
    \{thao\_nguyen3, nakul\_gopalan, stefanie\_tellex, george\_konidaris\}@brown.edu 
    \\ * Authors contributed equally
}



%

\maketitle

    \begin{abstract}
We develop a system to disambiguate object instances within the same class based on simple physical descriptions. The system takes as input a natural language phrase and a depth image containing a segmented object and predicts how similar the observed object is to the object described by the phrase. Our system is designed to learn from only a small amount of human-labeled language data and generalize to viewpoints not represented in the language-annotated depth image training set. By decoupling 3D shape representation from language representation, this method is able to ground language to novel objects using a small amount of language-annotated depth-data and a larger corpus of unlabeled 3D object meshes, even when these objects are partially observed from unusual viewpoints. Our system is able to disambiguate between novel objects, observed via depth images, based on natural language descriptions. Our method also enables view-point transfer; trained on human-annotated data on a small set of depth images captured from frontal viewpoints, our system successfully predicted object attributes from rear views despite having no such depth images in its training set. Finally, we demonstrate our approach on a Baxter robot, enabling it to pick specific objects based on human-provided natural language descriptions.

\end{abstract}

\IEEEpeerreviewmaketitle

\section{Introduction}
As robots grow increasingly capable of understanding and interacting with objects in their environments, a key bottleneck to widespread robot deployment in human-centric environments is the ability for non-experts to communicate with robots. One of the most sought after communication modalities is natural language, allowing a non-expert user to verbally issue directives. We focus our work here, applying natural language to the task of object-specification---describing which of many objects is being referred to by a user, is critically important when tasking a robot to perform actions such as picking, placing, or retrieving an item. Even simple commands, such as \textit{``bring me a small espresso cup"}, require object-specification to be successful.

Object-specification becomes far more difficult when the robot is forced to disambiguate between multiple objects of the same type (such as between several different teapots, rather than between a teapot and a bowl). The solution to this problem is to include descriptive language to not only specify object types, but also object attributes such as shape, size, or more abstract features. Many such descriptions are largely physical in nature, relating directly or indirectly to the actual shape of an object. For instance, descriptions like \say{round}, \say{tall}, \say{vintage}, and \say{retro}, all relate to shape. While current work exists to ground images to natural language \cite{krishnamurthy2013jointly, hu2016natural}, these systems only work when an object is observed from a similar viewpoint to human-annotated training images. We must also handle partially observed objects; it is unreasonable for a robot to observe objects in its environment from many angles with no self-occlusion. Furthermore, as human-annotated language data is expensive to gather, it is important to not require language annotated depth images from all possible viewpoints. A robust system must be able to generalize from a small human-labeled set of (depth) images to novel and unique object views.
\begin{figure}[t]
  \includegraphics[width=\linewidth]{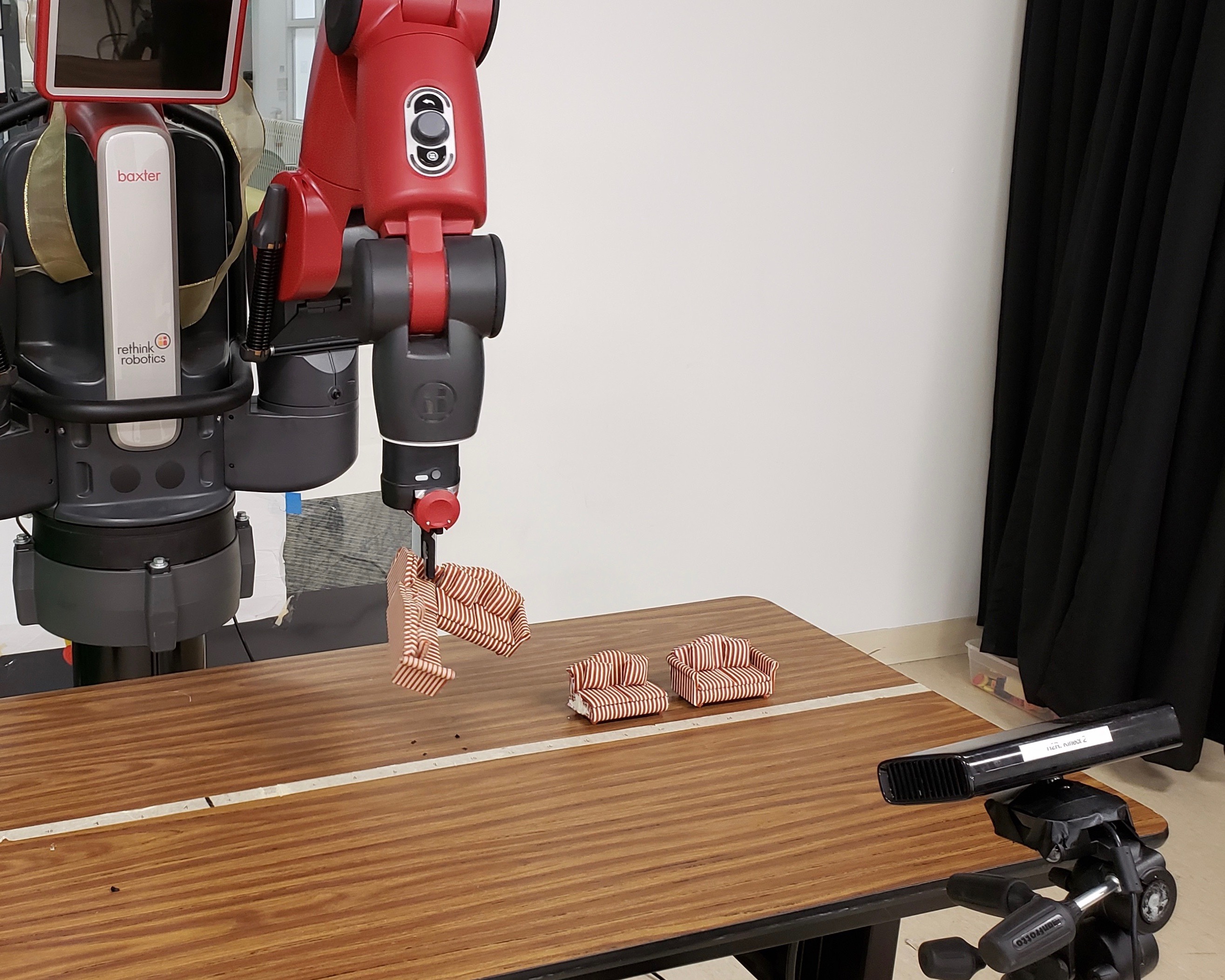}
  \caption{Our pipeline receives depth images and a natural language command such as, ``Pick up the bent couch", to retrieve the described couch.}
  \label{fig:coverImage}
\end{figure}
We develop a system that addresses these challenges by grounding descriptions to objects via explicit reasoning over 3D structure. We combine Bayesian Eigenobjects (BEOs) \cite{burchfiel2017bayesian}, a framework which enables estimation of a novel object's full 3D geometry from a single partial view, with a language grounding model. This combined model uses BEOs to predict a low-dimensional object embedding into a pose-invariant learned object space and predicts language grounding from this low-dimensional space. Critically, BEOs are well suited to the novel object case; using previously seen objects they generalize well to make predictions about previously unseen objects, even from partial views.

This structure has several advantages. First, it enables training the object-understanding portion of the system from a large set of non-language-annotated objects, reducing the need for expensive human-generated object attribute labels to be obtained for all the training data. Second, because the language model learns to predict language groundings from a low-dimensional subspace instead of high-dimensional 2.5D or 3D input, the model complexity, and amount of labeled training data required, is small; in our experiments, a simple 3-layer network was sufficient. Finally, because BEOs are already well suited to novel objects, our system scales well to new objects and partial views, an important capability for a robot operating in an unstructured human-centric environment. Unlike a single monolithic system which would require human-annotated depth images from all possible viewpoints, our approach allows a small number of annotated depth images from a limited set of viewpoints to generalize to significantly novel partial views and novel objects.

We evaluate our system on a dataset of several thousand ShapeNet object instances~\cite{shapenet2015} across three object classes, paired with human-generated object descriptions obtained from Amazon Mechanical Turk (AMT). We show that not only is our system able to distinguish between objects of the same class, it can do so even when objects are only observed from partial views. In a second experiment, we train our language model with a set of depth images taken only from the front of objects and can successfully predict attributes given test depth images taken from the rear. This view-invariance is a key property afforded by our use of an explicitly learned 3D representation---traditional fully supervised approaches are not capable of handling this scenario. Using our system, we demonstrate a Baxter robot successfully determining which object to pick based on a Microsoft Kinect depth image of several candidate objects and a simple natural language description of the desired object as shown in Figure~\ref{fig:coverImage}. Our system is fast, with inference running in under $100$ms for a single language+depth-image query.

\section{Background and Related Work}
Object retrieval refers to the task of finding and recovering an object specified by a human user, generally using natural language. While it is an important problem within artificial intelligence, it is particularly germane to robots, which have the ability to physically interact with specified objects.

The computer vision and natural language grounding communities attempt to solve object retrieval by locating or \textit{grounding} the object specified in an image using natural language \cite{krishnamurthy2013jointly, hu2016natural}. 
Krishnamurthy and Kollar \cite{krishnamurthy2013jointly} use a dataset of segmented objects and their natural language descriptions to learn the grounding of words to objects in the image by exploiting repeated occurrences of segmented objects within images, along with their descriptions in natural language. Hu et al. \cite{hu2016natural} use a similar approach albeit using deep neural networks to avoid parsing and feature construction by hand. Chen et al. \cite{chen2018text2shape} learn joint embeddings of language descriptions and colored 3D objects. 
These approaches are trained on image data collected from internet sources and may differ from data observed on a robot's camera. Moreover, these approaches are either 2D, making them sensitive to viewpoint differences, or not applicable to partially observed objects, making them difficult to apply to a real robot. Another relevant line of work is the SUN scene attribute dataset \cite{patterson2012sun} which maps images to attributes such as ``hills'', ``houses'', ``bicycle racks'', ``sun'', etc. Such understanding of image attributes provides high level scene descriptions, for example ``human hiking in a rainy field''. However, these approaches are typically unsuited for direct robotics application as they require explicit examples of similar objects from many viewpoints, all annotated with language, which is an expensive labeling task. Indeed, the typical robotics approach involves creating a database of all objects in the robot's environment and retrieving the object that fits the human requester's description  \cite{BORE2017139, shridhar2018interactive, interact_picking18, whitney2017reducing}. 

 

Bore et al. \cite{BORE2017139} use a mobile robot to create a scene with segmented 3D objects, where an object can be retrieved on demand using a query object, thus bypassing language grounding entirely. Whitney et al. \cite{whitney2017reducing} identifies objects in a scene using their descriptions given by a human user; the system  uses Partially Observable Markov Decision Processes (POMDPs) to ask questions to disambiguate the correct object. The system itself does not perform any classification, instead it uses POMDPs to choose the correct object after it is confident of the human's intention. Such an interactive language grounding has been improved by approaches such as Shridhar and Hsu \cite{shridhar2018interactive} and Hatori et. al \cite{interact_picking18}. They use a corpus of natural language question answering along with images of the objects in the robot's view to learn object grounding with questions and answers without a POMDP, by classifying the best fitting object and asking questions based on a handcrafted policy. The improvements here are that the grounding occurs in sensor data observed by the robot without any handcrafted features. However, the agents in these methods have seen all instances within the object set. These systems would have a hard time grasping or identifying a novel object given just a partial view, which is a common in robotics.

Separately from language grounding, recent years have seen an uptick of interest in the computer vision community on the problem of partial object completion. These methods take a partially observed object model (either in voxelized form or via a depth image) and attempt to predict an entire 3D model for the observed object. The most common of these approaches are \textit{object-database} methods which construct an explicit database of complete object scans and treat novel objects as queries into the database \cite{guided3DScanning, guibas_database_objects, guibas_database_objects2}. While these approaches can be successful if the query is very similar to a previously encountered object, they tend to have issues if objects are sufficiently novel. 
More recently, work has progressed on learned models that predict 3D shape from partially-observed input \cite{wu20153d, dai2017complete, Soltani2017Synthesizing3S, burchfiel2017bayesian}. While most of these methods require known voxelization of the partially observed query object, recent work has allowed for prediction directly from a depth image, an important requirement for most robotic applications \cite{burchfiel2018hbeo}.

Bayesian Eigenobjects (BEOs) \cite{burchfiel2017bayesian, burchfiel2018hbeo} offer compact representations of objects. Using 3D object models for training, BEOs generate a low-dimensional subspace serving as a basis which well captures the object classes used to train it. This decomposition is achieved via Variational Bayesian Principle Component Analysis (VBPCA) \cite{bishop1999variational}, ensuring that the learned subspace does not over-fit the training data. BEOs enable a number of useful features for robot perception and manipulation tasks, namely accurate object classification, completion, and pose estimation from object depth-map data. This makes the BEO subspace a compelling target for language mapping in a robotics domain. Critically, the hybrid variant of BEOs (HBEOs) \cite{burchfiel2018hbeo} learn an explicit subspace like BEOs, but use a deep convolutional network to learn an embedding directly from a depth image into the object-subspace, allowing for high performance, fast runtime, and the ability to complete objects without requiring voxelization of observed objects.




\section{Methods}
Our objective is to disambiguate between objects based on depth images and natural language descriptions. The naive approach would be to directly predict an object depth image, or its representation, given the object's natural language description. However, this approach is difficult to train, as depth-data is high-dimensional and highly continuous in nature. Even more critically, such an approach would require language+depth-image pairs with a large amount of viewpoint coverage, an unreasonable task given the difficulty of collecting rich human-annotated descriptions of objects. Instead, our approach factors the representation, allowing it to learn to reason about 3D structure from non-annotated 3D models, to learn a viewpoint-invariant representation, BEO, of object shape. We combine this representation with a small set of language data to enable object-language reasoning.

Given a natural language phrase and a segmented depth image, our system maps the depth image into a compact viewpoint-invariant object representation and then produces a joint embedding: both the phrase and the object representation are embedded into a shared low-dimensional space. Language descriptions and depth images that correspond will be close in this space, while pairs of depth images and descriptions that do not correspond will be relatively far apart.

We achieve this relationship by forcing the low dimensional representations of a given object's depth-data and its referring natural language data to align closely with each other by reducing a similarity metric during training. At inference time, we compute this similarity metric between a natural language command and candidate objects, observed via a depth image, in the joint low-dimensional embedding. We then select the object that has the most similar depth-data to the given natural language command within our learned embedding.
Our approach also allows the prediction of object attributes based on a depth image of an object or a natural language description of that object.

\begin{figure}[tb]
  \includegraphics[width=\linewidth]{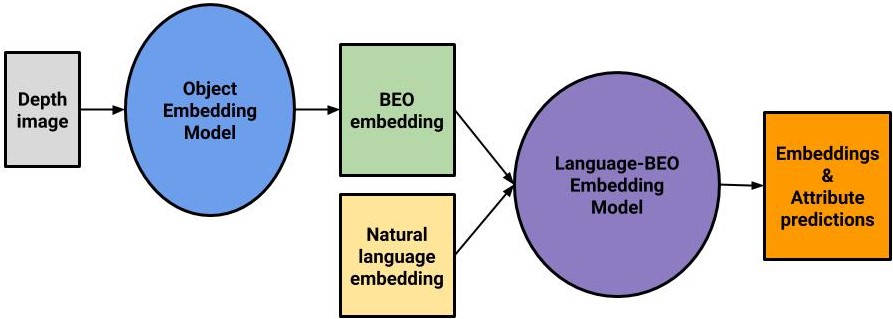}
  \caption{Diagram of the language grounding pipeline, consisting of the object and language-BEO embedding models.}
  \label{fig:pipeline}
\end{figure}

The language grounding system (shown in Figure \ref{fig:pipeline}) consists of two primary portions:  An object embedding portion (BEO module) which maps a partially observed object to a point in low-dimensional embedding space, and a language-BEO embedding portion which maps these embedded objects into attribute-based language descriptions. The two portions are trained in two different phases; first the BEO module is trained from a set of non-annotated 3D object models, and then the second phase learns a joint embedding of the natural language descriptions of the objects  and their corresponding BEO representations. The training data used for these two phases can be completely disjoint; there is no requirement that all objects used to train the BEO module are also used to train the language mapping portion. This permits a larger corpus of 3D shape data to be used with a small set of human-annotated data. Because the BEO module, once trained, can produce predictions from a single depth image, the human annotations need not be over entire 3D objects, but could instead be assigned to individual images for which no true 3D model is known. These annotations need not cover the entire range of viewpoints a robot might encounter, and could be from only a very limited set of views, because the BEO module provides for generalization across viewpoints.

\subsection{Obtaining Low-Dimensional BEO Embeddings}
The low-dimensional BEO embedding is learned from a set of 3D meshes from a single class. These meshes are aligned and voxelized as in Burchfiel and Konidaris \cite{burchfiel2018hbeo}. Each voxelized object is then stacked into a vector-representation; an object is now a point in high-dimensional voxel space. We perform VBPCA on these object points to yield a low-dimensional subspace defined by a mean vector, $\boldsymbol{\mu}$ and basis matrix, $\mathbf{W}$. We find an orthnormal basis $\mathbf{W'} = orth([\mathbf{W}, \boldsymbol{\mu}])$ using singular value decomposition and, with slight abuse of notation, hereafter refer to $\mathbf{W'}$ as simply $\mathbf{W}$.
Given a new (fully observed) object $\mathbf{o}$, we can obtain its embedding $\mathbf{o}'$ in this space via
\begin{equation}
\mathbf{o}'=\mathbf{W}^T\mathbf{o},
\label{eq:project_complete}
\end{equation}
and any point in this space can be \textit{back-projected} to 3D voxel space via
\begin{equation}
\mathbf{\hat{o}}= \mathbf{W}\mathbf{o}'.
\label{eq:recons}
\end{equation}
We use the HBEONet network structure \cite{burchfiel2018hbeo} to then learn a deep model for partially-observable inference which, given a depth image, directly predicts an embedding into the object subspace. During training, HBEONet receives depth images (generated synthetically from the 3D models) and target object embeddings. It consists of several convolutional and fully connected layers \cite{burchfiel2018hbeo}. This network is then used to train the language grounding portion of the system.

\subsection{Combining BEO Representations and Language}
\label{sec:methods}
\begin{figure}[tb]
  \includegraphics[width=\linewidth]{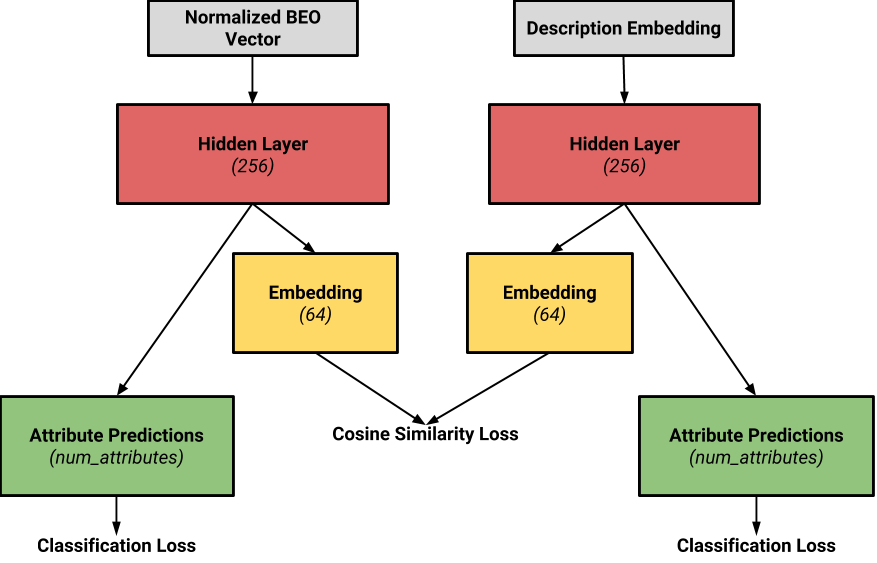}
  \caption{Diagram of the language-BEO embedding model. After training, the model outputs the cosine-similarity between a given language description and the BEO embedding of an object. The model can also predict attributes of an object given its BEO vector, or the object attributes given the natural language used to refer to the object. 
  \textit{num\_attributes} is number of attributes to be predicted.}
  \label{fig:modelDiagram}
\end{figure}

The BEO representations received from HBEONet are low-dimensional representations of object 3D meshes. Similarly, language can be represented in low-dimensional embeddings using techniques like Word2Vec \cite{mikolov2013efficient,mikolov2013distributed}, or GloVe~\cite{pennington2014glove}. However, the BEO embedding of an object has no connection to or similarity with the language embedding of the same object's natural language description. 
The BEO embeddings for a class of objects might be on a completely different manifold than the language embeddings for the descriptions of those objects, even if the dimensionality of the embedding spaces is the same. 
This discrepancy exists because these manifolds are learned for different tasks: BEO embeddings are learned for compactness of 3D object representation, whereas the language embeddings are learned for the predictive power of word representations in the context of a language model. 

To overcome this and learn a joint embedding space, we take an approach inspired by Siamese neural networks \cite{bromley1994signature}. We train a joint model which learns new embeddings for both BEO vectors and natural language. Embeddings of language descriptions and BEO vectors for similar objects should be close together in this new embedding space, whereas embeddings of BEO vectors and language  descriptions that do not correspond to similar objects should be farther apart. To measure closeness in this space, we use \textit{cosine similarity}:
\begin{equation}
    \textit{cosine\_similarity}(\pmb x, \pmb y) = \frac {\pmb x \cdot \pmb y}{\left| \pmb x \right| \cdot \left| \pmb y \right|}
\label{eq:cosim} 
\end{equation} 

The cosine similarity of two vectors is defined as the cosine of the angle between the vectors. It is calculated as the inner product of the vectors, normalized by the product of their lengths. The two vectors are maximally similar when their cosine similarity is $1$, and maximally dissimilar when their cosine similarity is $-1$.

Given an object's depth image and its corresponding language description, we first compute the low-dimensional BEO vector representation for the depth-data and then pass this representation to our joint embedding model shown in Figure~\ref{fig:modelDiagram}. This model takes as input a BEO vector and a bag-of-words sentence embedding of the natural language description. For sentence embeddings we use the mean GloVe \cite{pennington2014glove} embedding of all the words in a sentence. GloVe embeddings have been shown to be a robust sentence level feature \cite{wieting2015towards}.
Both the BEO vector and the sentence embedding are passed through hidden layers of size $256$, and the embedding dimension is then further reduced to size $64$. The cosine similarity loss is calculated between the $64$ dimensional embeddings of language and BEO vectors. 
We also predict object attributes directly from the $256$ dimensional hidden layers for language and BEO vectors (separately), which we train via binary cross-entropy loss.
As a result, for each of the inputs (natural language description and depth image), two outputs are produced: a $64$ dimensional joint embedding, and a prediction over the six attributes selected for each object class.\footnote{Each input modality, language and depth, takes an independent path through the network. Therefore, we can obtain our 64-dimensional embeddings and attribute predictions from just a BEO vector or language vector, which is required during attribute classification.}

\begin{table*}[h]
\centering
\caption{Object Retrieval Results}
\label{table:results}
\begin{tabular}{*{10}c}
\toprule
Object Class & 
Human Baseline &
\multicolumn{2}{c}{Full-view} & \multicolumn{2}{c}{Partial-view} & \multicolumn{2}{c}{View-transfer} \\

{}  & \textbf{\textit{Top-1 (Std. Error)}} & \textbf{\textit{Top-1}}   & \textbf{\textit{Top-2}}  & 
\textbf{\textit{Top-1}}   & \textbf{\textit{Top-2}} & \textbf{\textit{Top-1}}   & \textbf{\textit{Top-2}}\\
\midrule
Couch     & 77.3\% (0.025) & 60.6\%  & 87.1\%     & 58.6\%  & 86.3\% &  57.6\%  & 85.1\%\\
Car     & 76.0\% (0.024) & 74.6\%  & 93.8\%   & 73.5\%  & 93.4\%    & 72.2\%  & 93.1\%\\
Airplane     & 69.3\% (0.027) & 66.7\%  & 92.8\%    & 67.0\%  & 92.5\%   & 68.3\%  & 92.7\%\\
\bottomrule
\end{tabular}
\end{table*}

\subsection{Data Collection and ShapeNet Object Meshes}
\label{dataCollection}
We used three classes of objects from ShapeNet as our primary dataset: couches, cars, and airplanes. These models consist of 3D meshes, aligned and scaled to relatively consistent size. ShapeNet is a fairly noisy dataset with significant amounts of variation; it is not unusual to find models with extremely unusual shapes and some models may be incomplete. We manually removed objects from each category that were incorrectly classified or had significant damage to their meshes, but retained odd and unusual examples. During BEO training, each object was voxelized to a resolution of $64 \times 64 \times 64$ and $300$ dimensional object-subspaces were used, capturing between $80\%$ \textemdash \ $90\%$ of training object variance.
\begin{figure}[tb]
  \includegraphics[width=\linewidth]{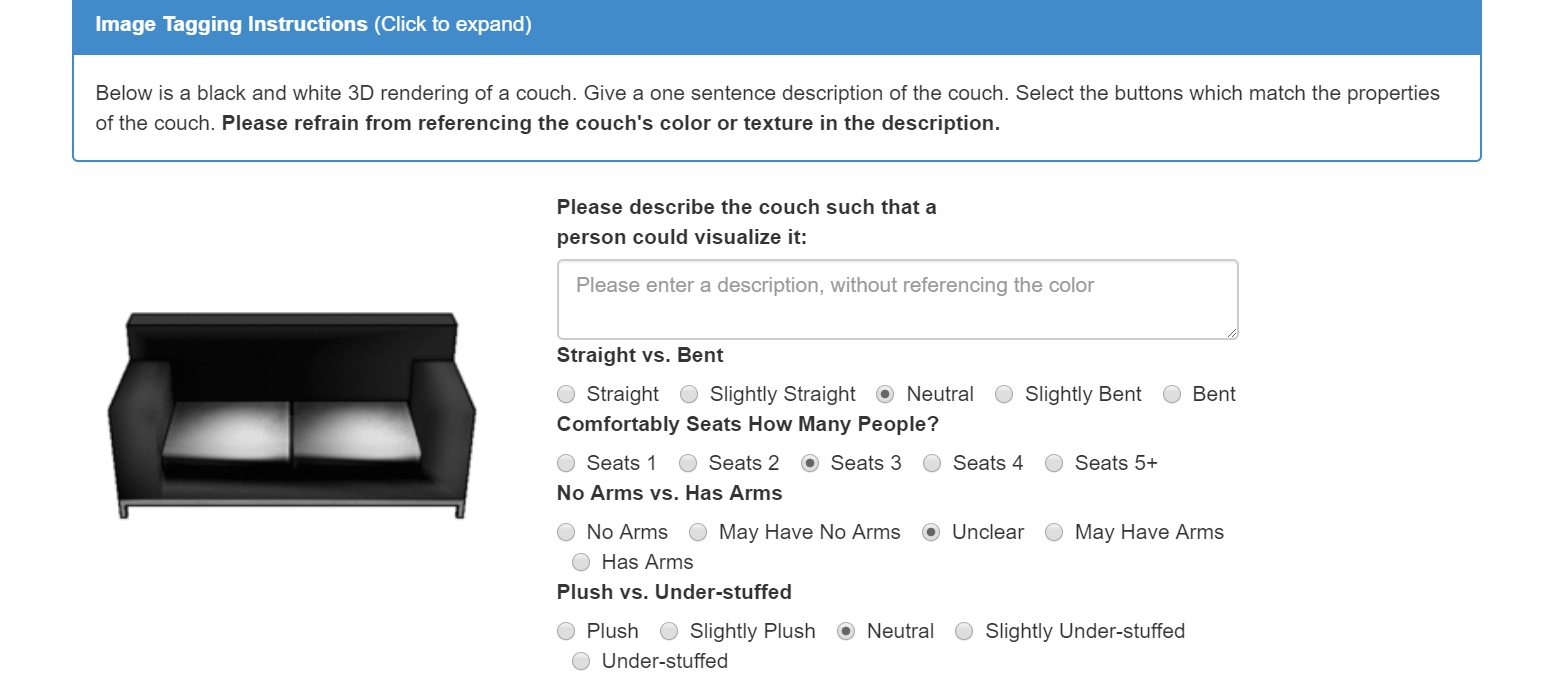}
  \caption{Amazon Mechanical Turk data collection interface. Workers were shown full 360-degree views of the 3D objects.}
  \label{fig:mturk}
\end{figure}

We collected natural language text descriptions and attribute labels, for the purpose of discriminating between different objects using language descriptions, through Amazon Mechanical Turk (AMT). The data collection interface can be seen in Figure \ref{fig:mturk}. AMT workers were shown full $360^{\circ}$ views of the 3D object instances, and got to select from a 5-point scale for each attribute classification.
The attribute labels were collected to serve two purposes: First, they allow us to observe the amount of object attribute signal present in the BEO representation. Second, we wanted to guide the natural language descriptions provided by the AMT workers, as otherwise the workers are likely to just describe an object with its name and not provide any object discriminative language. 
The attributes were chosen such that they describe physical and abstract properties of an object. For example, physical attributes such as long/short and abstract attributes such as luxury/economical were chosen for cars. 
We selected attributes which had the highest inter-rater agreement, as we wanted the attribute selections to have the highest consistency among English speakers. The chosen object attributes are listed in Table~\ref{table:attributes}.
\phantomsection
\label{para:data_desc}
AMT data were collected for $1250$ couch, $3405$ car, and $4044$ plane object instances. We collected $10$ sentence and attribute descriptions for each object instance.


\section{Experiments and Results}
Our evaluation was designed to verify that our combined model can accurately select an object given a natural language description of the object and depth image observations of possible candidate objects. We evaluate our model on two tasks: object retrieval and attribute classification. During object retrieval, the system observes three possible objects and must select the object corresponding to an input natural-language description. When performing attribute classification, the system observes a single object and must correctly predict its attributes.

We split our collected AMT data using $70\%$ for training, $15\%$ for development, and $15\%$ for testing.
Training examples for the language portion of the model consisted of pairs of depth-image-induced BEO object vectors  and corresponding language descriptions, along with the mean of the crowd-sourced attributes, binarized to signify the presence or absence of an attribute. Negative examples were generated by randomly selecting depth images and language descriptions that do not correspond. An equal number of positive and negative examples were shown to the network over the course of training. The cosine similarity loss we employed ensures that positive examples have higher cosine similarities and negative examples have lower cosine similarities in the $64$ dimensional space.\footnote{Training employed the Adam optimizer, with a learning rate of $0.0001$, and proceeded until accuracy on the development set began to decrease.} We then trained the language-object embedding of our model using $5$ randomly sampled depth images---and their resulting BEO shape vectors---from each object in the training set, along with the natural language description, and the binarized mean of human-provided attribute labels, corresponding to each depth image.

For the retrieval task, we show results for three training conditions: 1) \textit{Full-view}: a baseline where the system is given a ground-truth 3D model for the object it is observing, 2) \textit{Partial-view}: the scenario where the system is trained and evaluated with synthetic depth images over a wide variety of possible viewpoints for each object, 3) \textit{View-transfer}: the system is identical to the previous \textit{partial-view} case, except all training images come from a frontal viewpoint and all evaluation images are obtained from a side-rear view.
We train our model separately on all three object classes we sourced from ShapeNet. In all experiments, the input BEO embeddings to the language portion of our network were trained directly from all non-language-annotated meshes in the training dataset. In the partial-view and view-transfer 
cases, HBEONet was then trained using $600$ synthetically rendered depth images, across a variety of viewpoints, from each 3D mesh.
The baseline, full-view case, uses equation~\ref{eq:project_complete} instead of HBEONet to transform object observations into BEO embeddings.

To demonstrate the ability of our system to produce reasonable language attributes given a single query depth image, we also evaluate our model's attribute prediction performance against a human baseline.

Finally, to show the applicability of our approach to a real robot, we tested our system on a Baxter robot and demonstrated successful discrimination between three candidate objects given natural language descriptions and a Microsoft Kinect-generated depth image.

\label{expers}

\subsection{Object Retrieval from Natural Language}
\begin{figure*}[htb]
\centering
\includegraphics[width= 1.0\textwidth]{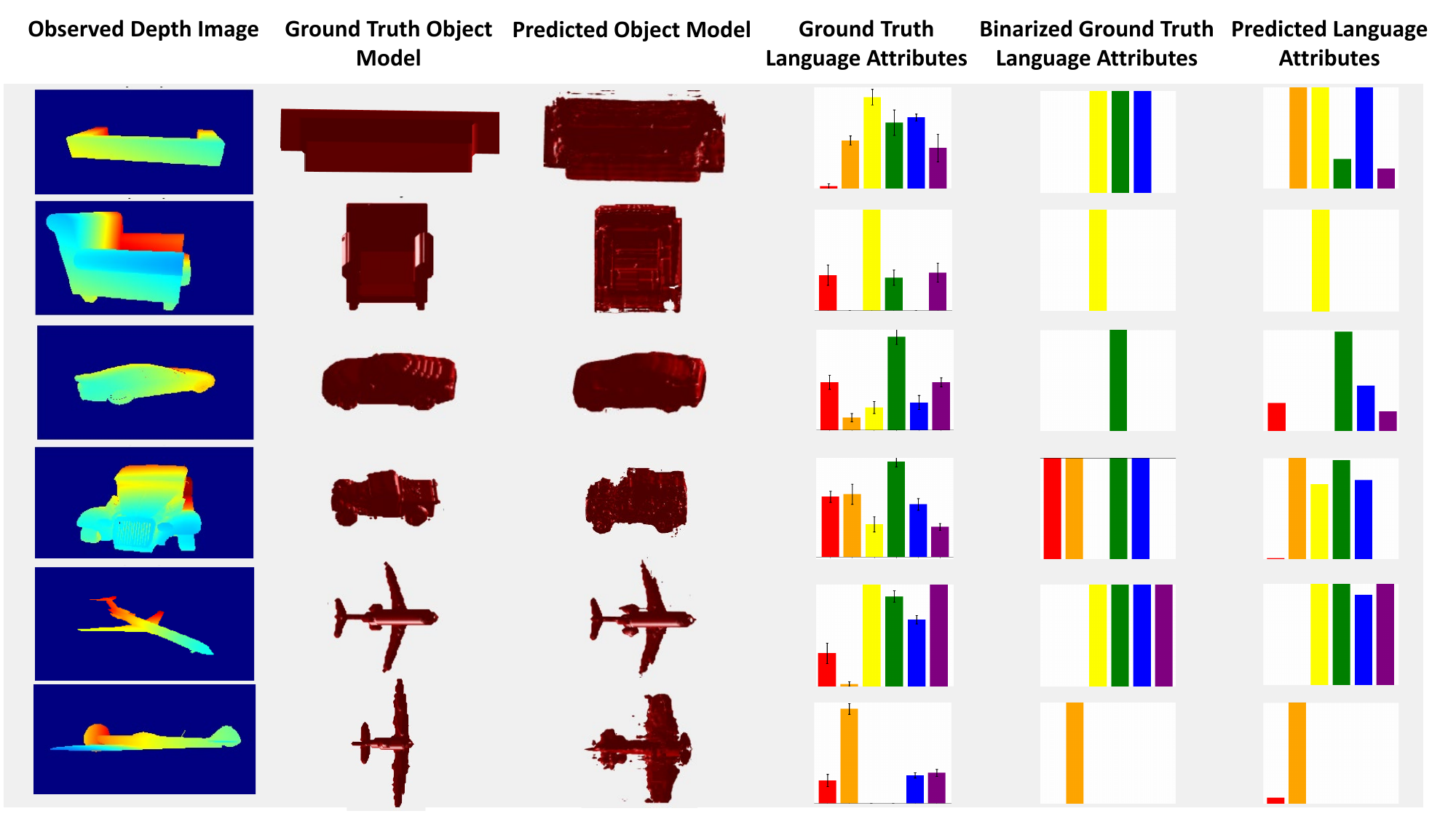}
\caption{Non-cherry-picked attribute prediction examples, produced by our model from single depth image input. The Predicted Object Model column visualizes BEO-predicted embeddings projected back into voxel space via equation \ref{eq:recons}.}
\label{fig:example_predictions}
\end{figure*}

\begin{figure}[tb]
  \includegraphics[width=\linewidth]{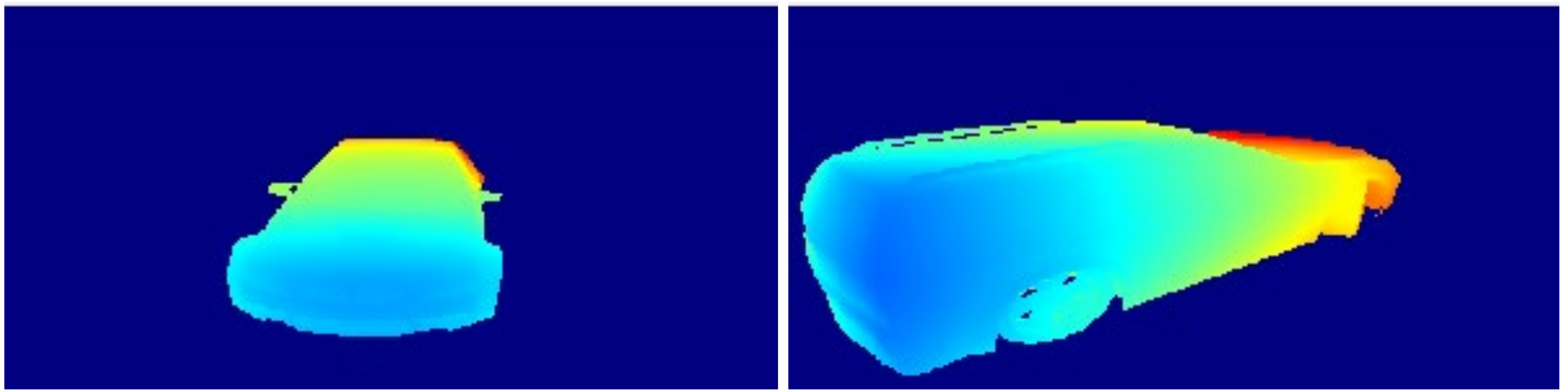}
  \caption{Example training depth image (left) and test depth image (right) from the view-transfer experiment. The underlying BEO representation maps these substantially different viewpoints to the same embedding space, helping the language grounding network ground object names and attributes to objects from unseen viewpoints.}
  \label{fig:view_transfer_example}
\end{figure}

To evaluate the retrieval performance of our model, we randomly selected a set of $10$ depth images for each object in our test set. In the full-view case, we simply used the 3D model for each object in the test set.

In each retrieval test, we showed the system three different objects along with a natural language description of one of those objects. We report the accuracy of our model in selecting the correct object in Table \ref{table:results} for both top-1 and top-2 retrieval metrics.

We also evaluated the robustness of our system to substantial viewpoint differences between testing and training data by training our language model with only frontal views while evaluating model performance based only on rear-side views. Figure \ref{fig:view_transfer_example} shows example training and testing depth images from our view-transfer experiment for a car object instance. Because the BEO representation is viewpoint invariant, our system successfully performs this task despite the dramatic distribution change between the training and evaluation datasets; see View-transfer results in Table \ref{table:results}.

We found that not only did the performance of our system decline only slightly between the fully-observable-object (full-view) baseline and the depth image-based partial-view scenario, training our model with annotated images from a single viewpoint (view-transfer) also failed to dramatically reduce performance. These results constitute a strong indicator that our system is successfully achieving view invariance and generalization by relying on the larger set of unlabeled object meshes used to train the BEO embeddings.

\subsection{Human Retrieval Baseline}
We compare the performance of our system to a human baseline for the retrieval task. Humans are expert symbol grounders and are able to ground objects from incomplete descriptions rather well from an early age~\cite{rakoczy2005children}.
We showed human users (full 360-degree views of) three objects 
and one language description, where the language was collected from AMT for one of the objects shown, and asked them to pick the object to which the language refers. The sample size for this experiment is $300$ users, and the study was done on AMT. The results are shown in Table \ref{table:results}. We found human performance similar to our system's top-1 retrieval accuracy in the full-view case. Even human users are not perfect at this task, as ambiguity exists between objects and their descriptions in many retrieval scenarios; because the set of three objects to be retrieved are chosen at random, there are often multiple very similar objects that must be disambiguated, resulting in multiple reasonable predictions.

\subsection{Object Attribute Prediction}
\begin{table*}[h]
\centering
\caption{Chosen Object Attributes and Our System's Respective Attribute Prediction F1 Scores}
\label{table:attributes}
\begin{tabular}{*{7}c}
\toprule
\textit{Object Class} & {Attr. 1} & 
{Attr. 2}  & {Attr. 3} & {Attr. 4}  & {Attr. 5}   & {Attr. 6}\\
\midrule
Couch & Straight/Bent  & Seats 1-5+  & No-Arms/Has-Arms  & Plush/Understuffed & Short/Long & Simple/Ornate\\ 
 & \textbf{0.87} & \textbf{0.87} & \textbf{0.80} & \textbf{0.64} & \textbf{0.83} & \textbf{0.67}\\
Car & Long/Short & Curvy/Boxy & Fast/Slow & Emergency/Not & Luxury/Economy & Seats 1-5+\\
 & \textbf{0.81} & \textbf{0.83} & \textbf{0.83} & \textbf{0.91} & \textbf{0.77} & \textbf{0.83}\\
Airplane &  Rounded/Angular & Commercial/Military & Propeller/Jet &  Historical/Modern & Engines 0-4+ & Seats 0-4+ \\
 & \textbf{0.88} & \textbf{0.91} & \textbf{0.94} & \textbf{0.93} & \textbf{0.73} & \textbf{0.88}\\
\bottomrule
\end{tabular}
\end{table*}
We also evaluated the 6-category object attribute predictions produced by our system; we predict language attributes for each of our object classes; Table~\ref{table:attributes} summarizes our system's performance on this task, while Figure~\ref{fig:example_predictions} illustrates several example attribute predictions produced by our system given a single query depth image. Attributes in each histogram are in the order, from left to right, that they are listed in Table~\ref{table:attributes}.
The first column of Figure~\ref{fig:example_predictions} illustrates the input depth image given to the system, the second column the observed object's ground truth shape, the third column the shape back-projected, via Equation~\ref{eq:recons} from the predicted BEO embedding, the fourth column contains raw attribute labels gathered from human-labelers for that object, the fifth column show these human-annotate labels after being binarized, and the sixth column shows the attribute predictions output by our system. Note that the system is trained with binarized attribute labels. The system was generally quite accurate when performing attribute predictions; it tends to be most confident in scenarios where human-labelers were highly mutually consistent, which likely resulted in increased consistency among attribute labels for similar objects in the training data.

\subsection{Picking Physical Objects from Depth Observations and Natural Language Descriptions}
We implemented our system on a Baxter robot: a mechanically compliant robot equipped with two parallel grippers. For this evaluation, we obtained realistic model couches (designed for use in doll houses) to serve as our test objects. We use the same (synthetically trained) network employed in the prior experiments, without retraining it explicitly on Kinect-generated depth images. We passed a textual language description of the requested object into our model along with a manually-segmented and scaled Kinect-captured depth image of each object in the scene. The robot then selected the observed object with the highest cosine-similarity with the language description and performed a pick action on it. Our system successfully picked up desired objects using phrases such as ``Pick up the couch with no arms." Despite having no doll-house couches or Kinect-produced depth images in its training data, our system was able to generalize to this new domain. 

  



\section{Future Directions}
\label{sec:future}
One of the critical challenges in robotics is scaling techniques to work in a wide array of environments. While the results presented in this paper constitute an important step towards grounding objects to natural language, existing work in this area still generally only applies to a small number, $1$ \textemdash \ $4$, of object classes~\cite{chen2018text2shape, yumer2015semantic}. Moving forward, more expansive datasets, of both natural language object descriptions and 3D shapes, are necessary to enable larger systems to be learned. While the ShapeNet object database is incredibly useful, it is still small by modern machine learning standards, containing roughly 50 classes (only 10 of which have at least 1k examples) and 50k object models. Just as the introduction of ImageNet, with a thousand classes and a million images, pushed the state of the art in 2D classification forward, a similarly sized shape dataset would be invaluable for 3D object understanding.

\section{Conclusion}
\label{sec:conclusion}
We developed a system to ground natural language descriptions of objects into a common space with observed object depth images. This scenario is challenging because novel objects will only be observed from a single, possibly unusual, viewpoint. Our approach decouples 3D shape understanding from language grounding by learning a generative 3D representation (using BEOs) that maps depth images into a viewpoint-invariant low-dimensional embedding. This 3D representation is trainable from non-annotated 3D object meshes, allowing us to label only a small subset of object depth images compared to the whole dataset. This viewpoint invariance and data efficiency is key due to the difficulty of acquiring large volumes of human-annotated labelling.

We show that our system successfully discriminates between objects given several segmented candidate depth images and a natural language description of a desired item, performing almost as well as a human baseline. This object discrimination is possible even in challenging scenarios where the training depth images and evaluation depth images are selected from entirely different viewpoints. We also demonstrate our method on a Baxter robot and show that the robot is able to successfully pick the correct item repeatedly based on a natural language description.

\section*{Acknowledgments}
We would like to acknowledge Edward Williams for his role in the conception of this project. This research was supported by the National Aeronautics and Space Administration under grant number NNX16AR61G, and in part by DARPA under agreement number D15AP00104. The U.S. Government is authorized to reproduce and distribute reprints for Governmental purposes notwithstanding any copyright notation thereon. The content is solely the responsibility of the authors and does not necessarily represent the official views of DARPA.

\bibliographystyle{IEEEtran}
{\tiny
\bibliography{IEEEabrv,references}
}
\end{document}